\documentclass[10pt,twocolumn,letterpaper]{article}

\usepackage[pagenumbers]{cvpr} 

\usepackage{makecell}
\usepackage{multirow}
\usepackage{booktabs}
\usepackage{amsfonts}
\usepackage{colortbl}

\usepackage{makecell}
\usepackage{multirow}
\usepackage{booktabs}
\usepackage{amsfonts}
\usepackage{colortbl}

\newcommand{\revise}[1]{{\color{black} #1}}
\newcommand{\dplus}[1]{\fontsize{6pt}{0.1em}\selectfont (\textbf{\textcolor{magenta}{#1}})}
\newcommand{\dminus}[1]{\fontsize{6pt}{0.1em}\selectfont (\textbf{\textcolor{cyan}{#1}})}

\definecolor{cvprblue}{rgb}{0.21,0.49,0.74}
\usepackage[pagebackref,breaklinks,colorlinks,allcolors=cvprblue]{hyperref}

\def\paperID{11737} 
\def\confName{CVPR}
\def\confYear{2025}

\title{SemiETS: Integrating Spatial and Content Consistencies for 
 \\ Semi-Supervised End-to-end Text Spotting}

\author{
Dongliang Luo$^*$\quad
Hanshen Zhu$^*$\quad
Ziyang Zhang\quad
Dingkang Liang\quad
Xudong Xie\quad \\
Yuliang Liu \quad
Xiang Bai$^\dagger$ \\
Huazhong University of Science and Technology\\
{\tt\small \{ldl, zhs\_china, zzyzz, dkliang, xdxie, ylliu, xbai\}@hust.edu.cn}
}

\begin{document}
\maketitle
\protect \renewcommand{\thefootnote}{\fnsymbol{footnote}}
\footnotetext[1]{Equal contribution. $^{\dag}$Corresponding author.} 

\begin{abstract}
  Most previous scene text spotting methods rely on high-quality manual annotations to achieve promising performance.
  To reduce their expensive costs, we study semi-supervised text spotting (SSTS) to exploit useful information from unlabeled images.
  However, directly applying existing semi-supervised methods of general scenes to SSTS will face new challenges: 1) inconsistent pseudo labels between detection and recognition tasks, and 2) sub-optimal supervisions caused by inconsistency between teacher/student.
  Thus, we propose a new \textbf{Semi}-supervised framework for \textbf{E}nd-to-end \textbf{T}ext \textbf{S}potting, namely \textbf{SemiETS} that leverages the complementarity of text detection and recognition.
  Specifically, it gradually generates reliable hierarchical pseudo labels for each task, thereby reducing noisy labels.
  Meanwhile, it extracts important information in locations and transcriptions from bidirectional flows to improve consistency.
  Extensive experiments on three datasets under various settings demonstrate the effectiveness of SemiETS on arbitrary-shaped text.
  For example, it outperforms previous state-of-the-art SSL methods by a large margin on end-to-end spotting (+8.7\%, +5.6\%, and +2.6\% H-mean under 0.5\%, 1\%, and 2\% labeled data settings on Total-Text, respectively).
  More importantly, it still improves upon a strongly supervised text spotter trained with plenty of labeled data by 2.0\%. Compelling domain adaptation ability shows practical potential.
  Moreover, our method demonstrates consistent improvement on different text spotters.
  Code will be available at \url{https://github.com/DrLuo/SemiETS}.
\end{abstract}

\section{Introduction}
\label{sec:intro}

\begin{figure}[tb]
  \centering
  \includegraphics[width=0.95\linewidth]{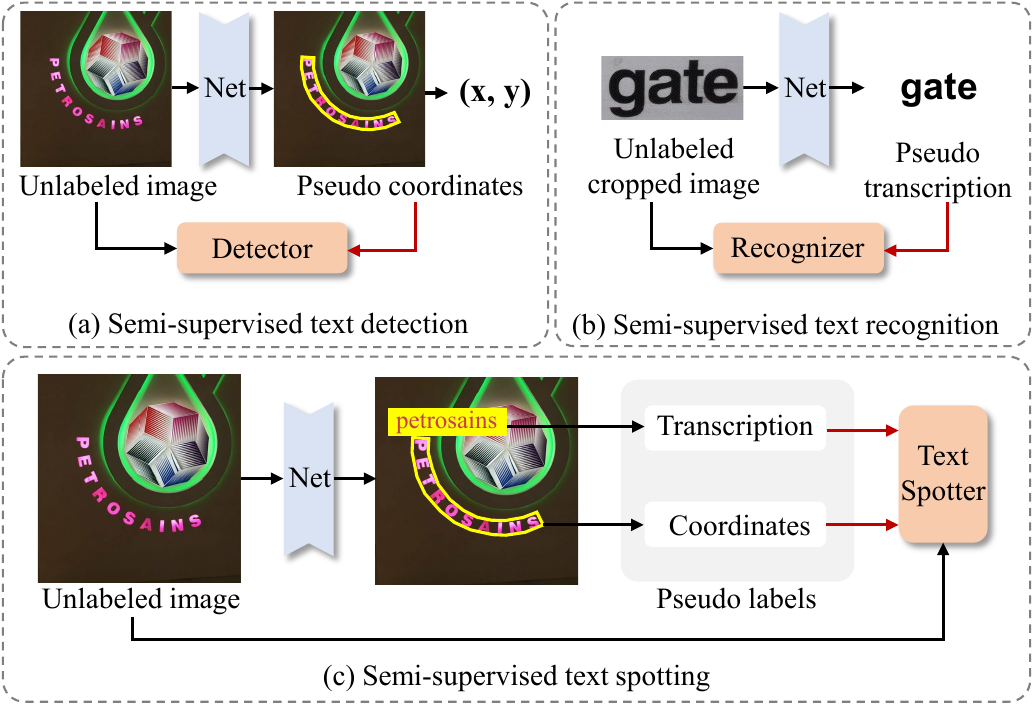}
  \vspace{-5pt}
  \caption{Comparison of different semi-supervised tasks related to text, including (a) semi-supervised text detection, (b) semi-supervised text recognition, and (c) semi-supervised text spotting. The \textcolor{BrickRed}{red} arrow indicates the supervision flow of pseudo labels.}
  \label{fig:background}
  \vspace{-10pt}
\end{figure}

Text spotting aims at concurrently localizing and recognizing texts from images.
Most previous works~\cite{mask_textspotter_v1,abcnet_v1,testr,swintextspotter,deepsolo,estextspotter} highly rely on high-quality annotations for satisfactory performance.
However, annotating such data is expensive and laborious due to its complex formats.
To alleviate the cost of labeling, exploring effective information from unlabeled text images is a valuable research direction. To achieve that, semi-supervised learning (SSL) is a natural and effective approach.

Recently, SSL methods have made considerable progress in general scenarios such as semi-supervised object detection (SSOD)~\cite{ARSL,consistent,zhang2023semi,shehzadi2024sparse}. The common practice is to use a teacher model to generate pseudo labels on unlabeled sets, which act as the ground truth (GT) for the student model~\cite{stac,mean_teacher}. Thus, migrating them to optical character recognition (OCR) is natural and straightforward. Although semi-supervised methods have been applied to OCR tasks, most focus on one single task of text detection~\cite{WeText,CurvedText,SemiText} or recognition~\cite{Whatif,pushing,FineGrained}. The former is mostly the modification of SSOD frameworks to fit the irregular shape of texts~\cite{CurvedText} or improve label selection~\cite{xiehongtao,DBSCAN_THR}, while not involving the recognition part, as in Fig.~\ref{fig:background} (a). The latter, illustrated in Fig.~\ref{fig:background} (b), usually performs pseudo-labeling~\cite{FineGrained, Seq-Ups} or consistency-based regularization~\cite{pushing,semi_yang} on cropped text regions, assuming the detection is already accurate beforehand. Differently, in semi-supervised text spotting (SSTS), both the location and content of text need to be carefully considered, as in Fig.~\ref{fig:background} (c), increasing the complexity. Not only the optimization goal is different, but also the relationship between detection and recognition is vital, hence bringing new challenges to SSL methods: 1) inconsistency between tasks that the pseudo labels of detection and recognition are not always accurate concurrently; 2) ambiguous and ineffective supervision signals caused by inconsistency between teacher and student.

Therefore, we propose a new framework for \textbf{Semi}-supervised \textbf{E}nd-to-end \textbf{T}ext \textbf{S}potting named \textbf{SemiETS} following the teacher-student architecture with two key designs: a Progressive Sample Assignment (PSA) module to tackle the inconsistency between tasks, and a Mutual Mining Strategy (MMS) to boost the effective supervision signals and reduce ambiguity. Specifically, PSA selects reliable pseudo labels by comprehensively considering the joint constraints and then gradually assigns hierarchical labels to each task. The quality of pseudo labels and the rationality of label assignment is thereby improved.

Furthermore, leveraging the complementarity between text detection and recognition, we excavate effective information using the proposed MMS to relieve the inconsistency problem from bidirectional flows. 
On the one hand, it estimates the reliability of transcription pseudo labels using the localization discrepancy between teacher and student. On the other hand, it propagates recognition information to the detection flow to softly amplify informative detection supervision signals to guide the student.

We conduct extensive experiments on a variety of scene text datasets. Experimental results demonstrate consistent superior performances of SemiETS over the state-of-the-art (SOTA) semi-supervised frameworks on arbitrary-shaped texts under various settings. Particularly, the improvements are more pronounced when using smaller proportions of labeled data.
For example, We significantly outperform the SOTA SSL methods~\cite{stac,mean_teacher} by 8.7\% and 4.7\% without lexicon on Total-Text and CTW1500 using only 0.5\% labeled data, respectively. 
Furthermore, it still improves upon a strong supervised baseline already trained using extensive labeled data with \revise{2.0\% H-mean in E2E (None)} on Total-Text, which verifies the potential of our method to further boost the ability of text spotters utilizing unlabeled data. Moreover, we prove SemiETS is compatible with various text spotters and brings consistent improvements.

The advantages of this paper can be summarized as:
1) We propose a semi-supervised framework, namely SemiETS, for end-to-end text spotting that can mine effective information from extensive unlabeled data. To our knowledge, this is the first effort exploring this task.
2) Aiming at the problem of inconsistency in semi-supervised text spotting, SemiETS consists of a Progressive Sample Assignment module to improve label reliability using hierarchy and a Mutual Mining Strategy to relieve ambiguity and amplify effective supervision signals.
3) SemiETS not only significantly outperforms existing SSL methods under various settings but can continue to boost performance upon a well-trained text spotter. Especially, it outperforms previous SSL SOTA by 8.7\% when only using 0.5\% labeled data.

\section{Related Work}

\subsection{End-to-End Text Spotting}

Text spotting aiming at concurrently localizing and recognizing texts concurrently~\cite {li2017towards}. Most studies mainly address two challenges: representing arbitrary shapes texts~\cite{mask_textspotter_v1,mask_textspotter_v2,mask_textspotter_v3,mango,pan++,boundary,abcnet_v1,abcnet_v2} and enhancing the intrinsic synergy of text detection and recognition~\cite{testr,deepsolo,swintextspotter,estextspotter}.
However, these methods require expensive labeling costs.
To reduce label dependency, SPTS series~\cite{spts_v1,spts_v2} simplify the spatial representation to a single point. Some~\cite{csvt,tts,wu2025wecromcl} even remove detection labels and only use transcriptions. However, at least the recognition GT is given to these methods. Thus, they cannot utilize information from unlabeled data. We explore the semi-supervised paradigm, focusing on generating high-quality pseudo labels for text spotting.

\subsection{Semi-supervised Object Detection}

It is a common practice for SSOD to generate pseudo labels using a teacher model and expect the student detectors to make consistent predictions on augmented input images.
STAC~\cite{stac} first trains a teacher detector with labeled data. 
Subsequent studies simultaneously update the teacher by EMA inherited from Mean-Teacher~\cite{mean_teacher}. Researchers are committed to improving the quality of pseudo labels for classification and regression in one-stage~\cite{unbiasedv2, zhou2022dense, consistent, ARSL}, two-stage~\cite{unbiased,xu2021end,tang2021humble,pseco} and DETR-based~\cite{zhang2023semi,shehzadi2024sparse} detectors, or extending to oriented objects~\cite{hua2023sood}.
However, SSOD methods for general scenarios hardly consider the characteristics of the text. Moreover, SSTS requires both text detection and recognition predictions, increasing its complexity.

\subsection{Semi-supervised Text Detection \& Recognition}

Most semi-supervised methods for OCR are designed only for text detection or recognition, while frameworks for text spotting are rarely studied.

\noindent\textbf{Semi-supervised Text Detection.}
WeText~\cite{WeText} proposes a semi-supervised framework for character detection, later extended to curved texts by Qin~\etal~\cite{CurvedText}. Subsequent methods refine pseudo-labeling by improving label quality. Some explore thresholding techniques~\cite{xiehongtao} or apply clustering~\cite{DBSCAN_THR}, while others use character-level information and context refinement to suppress false positives~\cite{SemiText}. However, similar to SSOD methods, these approaches seldom involve the recognition aspect.

\noindent\textbf{Semi-supervised Text Recognition.}
Baek~\etal~\cite{Whatif} apply pseudo-labeling to improve scene text recognition.
Due to the sequential property of the text, subsequent approaches aim to develop pseudo-labeling~\cite{FineGrained, Seq-Ups} and regularization~\cite{Gao, semi_yang, pushing} strategies for sequence. To select reliable pseudo labels, Li~\etal~\cite{FineGrained} set dynamic thresholds per character, while Seq-UPS~\cite{Seq-Ups} uses sequential uncertainty estimation. In regularization, some methods~\cite{Gao, semi_yang} apply sequence-level consistency, while Others use character-level regularization and tackle misalignment of characters by sharing context information~\cite{pushing} or reinforcement learning~\cite{semi_yang}. However, the input images for them are already focused on text regions, which is not guaranteed in SSTS, where detected regions might deviate.

Unlike the above works, in this paper, we explore semi-supervised text spotting, which reduces the annotation cost and boosts text spotters. In particular, we carefully consider the detection and recognition tasks concurrently.

\section{Preliminary}

\subsection{Task Definition}

Given a labeled image set $D_l=\{x_i^s, y_i^s \}_{i=1}^{N_l}$ and an unlabeled image set $D_u=\{x_i^u\}_{i=1}^{N_u}$, where $x_i^s$/$x_i^u$ represents labeled/unlabeled images and $N_l$/$N_u$ is the number of labeled/unlabeled images, respectively. Each label $y_i^s$ contains ground truth information for text spotting, including the coordinates $p$ and transcriptions $q$ of all text instances. Semi-supervised text spotting (SSTS) aims to leverage both labeled and unlabeled data to train a strong text spotter.

Following the pseudo-labeling paradigm inheriting from Mean-Teacher~\cite{mean_teacher}, our framework consists of a teacher model and a student model. The teacher generates pseudo labels from a weakly augmented image to guide the student, which takes a strongly augmented version as input. Additionally, the labeled dataset $D_l$ supervises the student in a standard manner. The teacher is updated simultaneously using Exponential Moving Average (EMA).

\subsection{Inconsistency Investigation}

In this part, we mainly summarize the inconsistent issues in semi-supervised text spotting from two aspects.

\begin{figure}[t]
  \centering
  \includegraphics[width=0.95\linewidth]{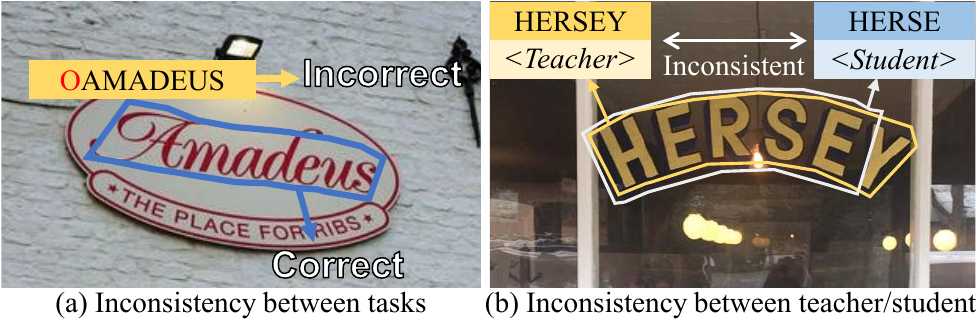}
   \vspace{-5pt}
  \caption{Illustration of the inconsistency issues including inconsistency (a) between tasks, and (b) between teacher and student.}
  \label{fig:challenge}
  \vspace{-10pt}
\end{figure}

\noindent\textbf{Inconsistency between tasks.} One challenge of SSTS is that the predictions contain two tasks, \ie, text detection and recognition. However, ensuring both pseudo labels are always reliable is not easy. For example, even though the location of the text region is precise, the recognition result may still be incorrect, as shown in Fig.~\ref{fig:challenge} (a). Naturally, simply using every predicted text instance equally to supervise the student is unsuitable, as it may introduce noisy labels.
Therefore, we propose a Progressive Sample Assignment module to distinguish reliable information.

\noindent\textbf{Inconsistency between teacher \& student.}
The teacher's and student's predictions may differ in text position or content. Since the recognition results are pretty sensitive to the detected regions, a slight deviation would cause a change in recognition results, as shown in Fig.~\ref{fig:challenge} (b). Intuitively, it is sub-optimal to directly force the student to learn the content of a misaligned region, which would lead to ambiguity.
Therefore, we propose a Mutual Mining Strategy to improve the alignment of recognition supervision and boost proper supervision signals to guide the student.

\section{Methodology}

\begin{figure*}[tb]
  \centering
  \includegraphics[width=0.90\linewidth]{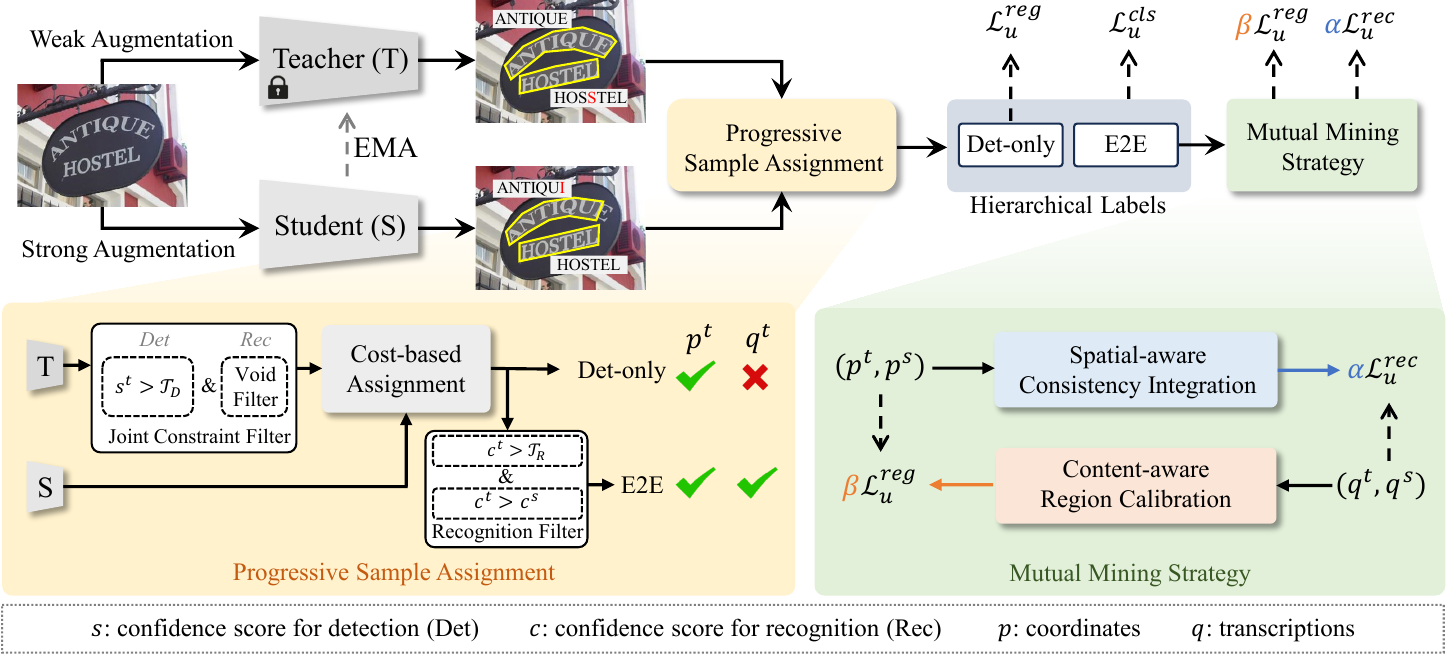}
  \vspace{-5pt}
  \caption{Overview of the proposed framework where the labeled data flow is omitted. Given an unlabeled image, Progressive Sample Assignment selects reliable pseudo labels and splits them into Det-only and E2E labels. Then, the Mutual Mining Strategy explores effective information in E2E labels in a crossover strategy with Spatial-aware Consistency Integration and Content-aware Region Calibration.}
  \label{fig:pipeline}
  \vspace{-10pt}
\end{figure*}

The overall framework of SemiETS is illustrated in Fig.~\ref{fig:pipeline} focusing on unlabeled data flow. It consists of two key components: A Progressive Sample Assignment (PSA) module improves the quality of pseudo labels to supervise the student; the Mutual Mining Strategy (MMS) further mines useful information from E2E labels to relieve inconsistency. We take DeepSolo~\cite{deepsolo}, a recent DETR-based text spotter with a concise architecture, as an example.

\subsection{Progressive Sample Assignment}

Generating high-quality pseudo labels is essential for semi-supervised learning but is challenging in SSTS.
Due to task-wise inconsistency, selecting pseudo labels solely from a single perspective is inadequate and rigid.
Thus, we propose PSA to progressively distinguish useful localization and recognition labels, constructing hierarchical supervisions to suppress noisy information.

As shown in Fig.~\ref{fig:pipeline}, the PSA consists of two steps. Firstly, a joint constraint filter first removes predictions with the classification score $s^t$ lower than a threshold $\mathcal{T}_D$ or with void decoded text.
Then, the student's predictions are matched to the remaining samples through the cost-based assignment. The cost function comprehensively integrates the detection and recognition, which is formulated as:

\begin{equation}
\begin{split}
    \mathcal{C}(\hat{Y}_i^s, Y_j^t) = &\lambda_{\text{cls}}\text{FL}(s_i^s, B(s_j^t)) + \lambda_{\text{text}}\mathcal{L}_{text}(\hat{q}^s_i, q^t_j) \\ &+ \lambda_{\text{coord}} \Vert p^s_i, p^t_j \Vert_1,~~ i \in N, j \in M,
\end{split}
\end{equation}

\noindent where $\hat{Y}_i^s$ is the predictions of student and $Y_j^t$ is the coarsely selected pseudo labels. FL$(*)$ is derived from the focal loss. $B$ is binarization. $\mathcal{L}_{text}$ is the recognition loss between the predicted characters $\hat{q}^s_i$ and the decoded transcription $q^t_j$. $p_i$ refers to the points on the polygon. $M$ and $N$ denote the number of student's predictions and pseudo labels, respectively. $\lambda_\text{cls}$, $\lambda_\text{text}$, and $\lambda_\text{coord}$ are set to 1.0, 1.0, and 0.5 by default, respectively. Hungarian algorithm is leveraged to establish optimal bipartite matching.
To improve the training efficiency in DETR-based text spotters, we introduce the hybrid matching (HM) strategy~\cite{zhang2023semi} that adopts the one-to-many assignment in early iterations. Here, each matched pseudo label is valid to supervise the detection flow.

In the second step, we further select reliable recognition labels with a recognition filter combining the following process:
1) filtering by threshold: the confidence score $c^t$ of the pseudo label is larger than a certain threshold $\mathcal{T}_R$;
2) confidence comparison (CC): $c^t$ is larger than the matched student's confidence score $c^s$.
CC is used to ensure the teacher has positive guidance for the student.

As a result, pseudo labels are categorized into Det-only and E2E. The transcriptions of Det-only labels are not included in the optimization. Unlike treating every instance equally, our differentiated assignment maximizes the utilization of data while reducing misleading information, thereby improving the quality of supervision for both tasks.

\begin{figure}[tb]
  \centering
  \includegraphics[width=0.78\linewidth]{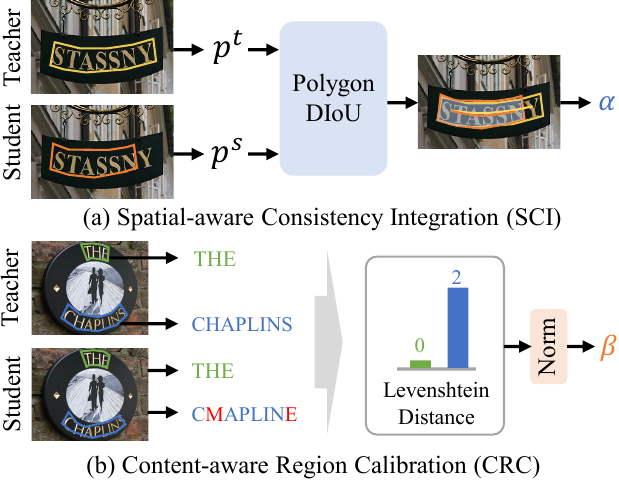}
  \vspace{-5pt}
  \caption{The details of SCI and CRC in the proposed Mutual Mining Strategy. We distinguish text instances using \textcolor{YellowGreen}{green} and \textcolor{RoyalBlue}{blue} in (b) and the wrongly recognized characters indicated in \textcolor{red}{red}.}
  \label{fig:modules}
  \vspace{-10pt}
\end{figure}

\subsection{Mutual Mining Strategy}

Since text detection and recognition have a natural synergy, we further develop MMS to mine useful information by explicitly bridging the correlation between these two tasks.
As illustrated in Fig.~\ref{fig:pipeline}, it consists of a Spatial-aware Consistency Integration (SCI) and a Content-aware Region Calibration (CRC) to conduct mutual mining in bidirectional flows. Specifically, SCI integrates the detection accuracy into the recognition flow. In turn, CRC facilitates the learning of detection incorporated with recognition information.

\subsubsection{Spatial-aware Consistency Integration}

Since text recognition results are sensitive to the detected text regions, the misaligned regions between teacher and student may cause ambiguous recognition supervision. Meanwhile, certain pseudo labels could be erroneous despite having high confidence scores.
Therefore, we propose SCI to integrate detection information into the recognition flow, as shown in Fig.~\ref{fig:modules} (a).
Firstly, softly adjusting the importance of the recognition supervision according to region alignment can relieve ambiguity, and enable the framework to better understand the geometric priors of recognition.
Secondly, the discrepancy in detected regions between teacher and student can be regarded as the uncertainty of pseudo labels, thereby estimating their reliability.

Considering that the scale and shape of text vary, and characters are usually distributed in order, we migrate Distance-IoU (DIoU)~\cite{diou} to polygon to measure spatial consistency rather than vanilla IoU. 
For the $i$-th matched pairs whose detected regions are represented by polygons $p_i^t$ and $p_i^s$, the polygon DIoU is formulated as:

\vspace{-7pt}
\begin{equation}
  \text{DIoU}(p_i^t, p_i^s) = \text{IoU}(p_i^t, p_i^s) - \frac{\frac{1}{K}\sum_{k=1}^{K}{E(\widetilde{p}^t_{(i,k)}, \widetilde{p}^s_{(i,k)})}}{\mathop{\text{max}}\limits_{1 \leq m, n \leq 2K}\  E(\widehat{p}_{(i,m)}^t, \widehat{p}_{(i,n)}^s)},
\end{equation}

\noindent where $\widetilde{p}_{(i,m)}$ and $\widehat{p}_{(i,n)}$ are the $m$-th center point and the $n$-th boundary point of the corresponding region. $K$ is the number of center points within a text instance and $E(*)$ means the Euclidean distance of two points. It integrates the deviations in center points, overlap between polygons, and text instance scale. Thus, we obtain the adaptive factor to adjust the recognition loss of the semi-supervised flow:

\begin{equation}
    \alpha_i = 1 + \text{DIoU}(p_i^t, p_i^s).
\end{equation}

\noindent Using the spatial-aware modulating factor, SCI suppresses ambiguous optimization, facilitating the learning process.

\subsubsection{Content-aware Region Calibration}

Conversely, the recognition results can help calibrate the predicted text regions.
We assume that the discrepancy between text contents predicted by the teacher and student implicitly indicates the deviation of the detected text regions. Therefore, we propose CRC to amplify important supervision signals for detection from recognition results as shown in Fig.~\ref{fig:modules} (b).
Such design lets the optimization focus on the imprecise regions that cause incorrect transcriptions in particular.
Specifically, we leverage Levenshtein distance to measure the disparity between words, formulated as:

\vspace{-5pt}
\begin{equation}
  D(q_i^t, q_i^s) = \frac{Levenshtein(q_i^t, q_i^s)}{\text{max}(|q_i^t|,|q_i^s|)},
\end{equation}

\noindent where $q_i^t$ and $q_i^s$ are the $i$-th pair of decoded words from teacher and student, respectively. The Levenshtein distance is normalized by the maximum length of words. Then, the corresponding factor $\beta_i$ for regression loss is derived as:

\begin{equation}
  \beta_i = 
  \begin{cases} 
  1 + \lambda D(q_i^t, q_i^s), & \text{if } c^t_i > \mathcal{T}_R ~\text{and}~ c^t_i > c^s_i, \\
  1,  & \text{otherwise},
  \end{cases}
\end{equation}

\noindent where $\lambda$ is a scale factor. CRC is only applied to E2E labels. The content-aware modulating factor highlights the inconsistent detection and explicitly enhances the task synergy.

\subsection{Training Objective}

The overall loss $\mathcal{L}$ of SemiETS is formulated as:

\vspace{-5pt}
\begin{equation}
  \mathcal{L} = \omega_l \mathcal{L}_l + \omega_u \mathcal{L}_u^{cls} + \omega_u ( \beta \cdot \mathcal{L}_u^{reg} + \alpha \cdot \mathcal{L}_u^{rec} ),
\end{equation}

\noindent where $\mathcal{L}_l$ is the supervised loss inherit from the supervised baseline. $\mathcal{L}_u^{cls}$, $\mathcal{L}_u^{reg}$ and $\mathcal{L}_u^{rec}$ are semi-supervised classification, regression and recognition losses in the unlabeled data flow, respectively.
$\omega_l$ and $\omega_u$ are the weighting factors.

\section{Experiments}

\subsection{Experimental Settings}

\textbf{Datasets.} We conduct experiments on three widely used scene text datasets, \ie, Total-Text~\cite{total}, ICDAR 2015~\cite{ic15} (IC15), and CTW1500~\cite{ctw1500}.

\noindent\textbf{Partially Labeled Data.} We randomly sample 0.5\%, 1\%, 2\%, 5\%, and 10\% images from the training set of each dataset as labeled data and set the remaining images as unlabeled data following the data split in SSOD~\cite{consistent,zhang2023semi}. The smallest labeled subset only contains 5 samples.

\noindent\textbf{Fully Labeled Data.} We set the training set of Total-Text or IC15 as labeled data and images from TextOCR~\cite{textocr} as additional unlabeled data.

\noindent\textbf{Domain Adaptation.} We set the training set of Total-Text and IC15 as the labeled and unlabeled data, respectively. Models are evaluated on the test set of IC15.

\subsection{Implementation Details}

We first pre-trained the text spotter on Synth150K~\cite{abcnet_v1} to initialize the model's text perception ability.
We use DeepSolo~\cite{deepsolo} as our base text spotter.
The labeled-to-unlabeled data ratio for Partially Labeled Data settings is 1:2 on Total-Text and CTW1500, and 1:1 on ICDAR 2015. AdamW is the optimizer. $\lambda$ is set to 20, and $\mathcal{T}_R$ is set to 0.7. $\omega_s$ and $\omega_u$ are set to 1 and 2, respectively. We use the data augmentation strategy from Semi-DETR~\cite{zhang2023semi}, excluding flipping and polarization to avoid ambiguity in recognized texts.

\subsection{Main Results}

\subsubsection{Partially Labeled Data}

\begin{table*}[tb]
\scriptsize
\setlength{\tabcolsep}{1.2mm}
  \caption{The end-to-end spotting results on curved text datasets Total-Text and CTW1500 under the Partially Labeled Data setting. None (Full) denotes the F1-measure without (with) using the lexicon that includes all words in the test set. Experiments are conducted on various data proportions. `*' indicates our adaptation of SSOD methods that fit curved texts and the recognition loss from pseudo labels is added.}
  \vspace{-8pt}
  \label{tab:curve_e2e}
  \centering
  \begin{tabular}{@{}lcccccccccc|cccccccccc@{}}
    \toprule
    \multirow{4}{*}{Methods} & \multicolumn{10}{c}{Total-Text} & \multicolumn{10}{c}{CTW1500} \\
    \cmidrule(lr){2-11}\cmidrule(lr){12-21}
     &  \multicolumn{2}{c}{0.5\%} &  \multicolumn{2}{c}{1\%} &   \multicolumn{2}{c}{2\%} &  \multicolumn{2}{c}{5\%} & \multicolumn{2}{c}{10\%} &  \multicolumn{2}{c}{0.5\%} &  \multicolumn{2}{c}{1\%} &   \multicolumn{2}{c}{2\%} &  \multicolumn{2}{c}{5\%} & \multicolumn{2}{c}{10\%} \\ 
    \cmidrule(lr){2-3}\cmidrule(lr){4-5}\cmidrule(lr){6-7}
    \cmidrule(lr){8-9}\cmidrule(lr){10-11}\cmidrule(lr){12-13}\cmidrule(lr){14-15}\cmidrule(lr){16-17}\cmidrule(lr){18-19}\cmidrule(lr){20-21}
     & None & Full & None & Full & None & Full & None & Full & None & Full & None & Full & None & Full & None & Full & None & Full & None & Full\\
     \midrule
   Supervised~\cite{deepsolo} & 58.8 & 71.1  & 61.2  & 74.9  & 63.4 & 76.8 & 66.9 & 78.8 & 69.9 & 80.9 & 30.1 & 59.4 & 40.6 & 66.2  & 48.1  & 71.2 & 53.0 & 74.8 & 56.9  & 76.9 \\
   \midrule
   STAC*~\cite{stac} & \underline{63.3}  & \underline{73.7}  & 66.7  & \underline{77.2}  & 67.3 & \underline{77.7} & 72.1 & \underline{80.9} & 73.2 & 82.8 & 32.7 & 60.7 & 43.2 & 68.0 & 49.7 & 72.0 & 55.2 & 76.3 & 58.4 & \underline{77.5} \\
   Mean-Teacher*~\cite{mean_teacher} & 55.5  & 57.7  & 64.4  & 69.0  & 68.6 & 73.7 & \underline{73.5} & 80.6 & 73.8 & 81.9 & \underline{48.8} & \underline{71.0} & \underline{52.7} & \underline{73.1} & \underline{55.0} & \underline{75.1} & 57.8 & \underline{77.1} & 58.1 & 77.3 \\
   Soft Teacher*~\cite{xu2021end} & 58.4 & 60.9 & 61.2 & 63.8 & 68.2 & 73.9 & 72.4 & 78.9 & 73.4 & 81.3 & 41.7 & 59.8 & 52.5 & 71.2 & 53.6 & 73.4 & \underline{58.5} & 76.2 & \underline{58.9} & 77.1  \\
   Unbiased Teacher v2*~\cite{unbiasedv2} & 56.6 & 58.9 & 64.7 & 69.5 & \underline{70.8} & 77.3 & 73.3 & 80.7 & \underline{75.0} & \underline{83.5} & 47.5 & 66.1 & 50.4 & 68.0 & 52.3 & 69.3 & 56.7 & 73.8 & 56.5 & 74.5 \\
   Semi-DETR*~\cite{zhang2023semi} & 59.4 & 63.0 & \underline{67.2} & 73.0 & 68.5 & 75.5 & 71.5 & 79.4 & 74.2 & 82.2 & 33.1 & 49.7 & 38.2 & 53.2 & 41.3 & 55.6 & 47.7 & 63.6 & 57.2 & 75.3  \\
   \rowcolor{gray!10}
   SemiETS (Ours) &\textbf{72.0}  &\textbf{78.5}  &\textbf{72.8}  &\textbf{80.6}  & \textbf{73.4} & \textbf{82.2} & \textbf{75.4} & \textbf{83.1} & \textbf{76.4}& \textbf{84.8} & \textbf{53.5} & \textbf{75.6} & \textbf{56.2} & \textbf{76.3} & \textbf{58.6} & \textbf{77.6} & \textbf{60.0} & \textbf{79.0} & \textbf{60.9} & \textbf{79.6} \\
  \bottomrule
  \end{tabular}
  
\end{table*}

\begin{table*}[tb!]
\scriptsize
\setlength{\tabcolsep}{2.5mm}
  \caption{The end-to-end spotting results on ICDAR 2015 under the Partially Labeled Data setting. ‘S’, ‘W’ and ‘G’ refer to using strong, weak and generic lexicons, respectively. Experiments are conducted on 0.5\%, 1\%, 2\%, 5\% and 10\% labeled data settings.}
  \vspace{-8pt}
  \label{tab:ic15_e2e}
  \centering
  \begin{tabular}{@{}lccc|ccc|ccc|ccc|ccc@{}}
    \toprule
    \makecell[l]{\multirow{2}{*}{Methods}} &  \multicolumn{3}{c}{0.5\%} &  \multicolumn{3}{c}{1\%} &   \multicolumn{3}{c}{2\%} &  \multicolumn{3}{c}{5\%} & \multicolumn{3}{c}{10\%} \\ 
    \cmidrule(lr){2-4}\cmidrule(lr){5-7}\cmidrule(lr){8-10}
    \cmidrule(lr){11-13}\cmidrule(lr){14-16}
     & S & W & G & S & W & G & S & W & G & S & W & G & S & W & G \\
     \midrule
   Supervised~\cite{deepsolo} & 71.2 & 66.1 & 59.9 & 69.9 & 65.6 & 59.8  & 72.0 &	67.4 &	61.6 & 77.3 & 72.5 & 67.3 & 79.3 & 75.0 & 69.2  \\
   \midrule 
   STAC*~\cite{stac} & 76.6 & 71.2 & 64.4 & 75.0 & 69.7 & 63.7 & 76.6 & 71.7 & 64.8 & 80.8 & 75.3 & \underline{69.6} & 81.2 & 76.8 & \underline{71.2} \\
   Mean-Teacher*~\cite{mean_teacher} & 77.7 & 67.9 & 60.6 & 76.8 & 68.5 & 61.9 & 78.6 & 70.5 & 63.8 & 81.6 & 74.0 & 67.9 & 82.3 & 76.0 & 70.2 \\
   Soft Teacher*~\cite{xu2021end} & 76.5 & 67.2 & 60.4 & 76.2  & 68.2  & 61.8  & 77.7 & 69.1 & 62.5 & 81.4 & 73.5 & 67.1 & 81.4 & 74.3 & 68.4 \\
   Unbiased Teacher v2*~\cite{unbiasedv2} & 74.9 & 64.7 & 57.5 & 75.4 & 66.3 & 59.2 & 77.2 & 69.1 & 61.5 & 79.6 & 71.7 & 65.6 & 80.7 & 74.0 & 67.5 \\
   Semi-DETR*~\cite{zhang2023semi} & \underline{79.9} & \underline{71.6} & \underline{65.3} & \underline{80.0} & \underline{72.6} & \underline{66.2} & \underline{81.2} & \underline{73.9} & \underline{67.5} & \underline{82.7} & \underline{75.5} & 69.3 & \underline{83.2} & \underline{77.1} & 71.1  \\
    \rowcolor{gray!10}
   SemiETS (Ours) & \textbf{81.9} & \textbf{74.5} & \textbf{67.8} & \textbf{82.5} & \textbf{75.3} & \textbf{68.8} & \textbf{82.6} & \textbf{76.2} & \textbf{70.0} & \textbf{84.0} & \textbf{77.4} & \textbf{71.2} & \textbf{84.1} & \textbf{77.9} & \textbf{71.8} \\

  \bottomrule
  \end{tabular}

\end{table*}

\noindent We compare our method to several popular semi-supervised learning methods, including STAC~\cite{stac}, Mean-Teacher~\cite{mean_teacher}, Soft Teacher~\cite{xu2021end}, Unbiased Teacher v2~\cite{unbiasedv2}, and Semi-DETR~\cite{zhang2023semi}.
We conduct the partially labeled data settings on three popular benchmarks for text spotting.
In general, our method achieves the best results on all of them. In addition, its advantage is more pronounced using less proportions of labeled data such as 0.5\%. 

\noindent\textbf{Results on Total-Text.}
As shown in Tab.~\ref{tab:curve_e2e}, our framework archives state-of-the-art end-to-end recognition results on arbitrary-shaped texts under all proportions. Especially, when only using 0.5\% labeled data, it outperforms the previous SOTA by 8.7\% without lexicon. Moreover, the supervised baseline can be improved by 13.2\%.
Surprisingly, the performance of most EMA-based methods~\cite{mean_teacher,xu2021end,unbiasedv2} even degrades in low data proportions. We assume it is because models with limited capacity are likely to introduce cumulative error without specified designs for SSTS.

\noindent\textbf{Results on ICDAR 2015.}
SemiETS also achieves the best results under all proportions on multi-oriented scene text as presented in Tab.~\ref{tab:ic15_e2e}, indicating its effectiveness in complex scenarios. 
Specifically, it surpasses the SOTA methods by 2.5\%, 2.6\%, and 2.5\% H-mean under 0.5\%, 1\%, and 2\% labeled data settings using generic lexicon, respectively.

\noindent\textbf{Results on CTW1500.}
The results in Tab.~\ref{tab:curve_e2e} also demonstrate the consistent superiority of SemiETS in spotting curved text at the text-line level. Even only using 5 labeled images, SemiETS achieves decent results, bringing 23.4\% performance gain to the supervised baseline in E2E (None).

\subsubsection{Fully Labeled Data}

Under this setting, we can explore the potential of the proposed semi-supervised framework even if it has been trained with extensive labeled data. As shown in Tab.~\ref{tab:full}, although the supervised baseline has already achieved a high performance of 79.7\% H-mean in E2E (None) on Total-Text, our SemiETS still takes 2\% improvements on it. In addition, our method performs better than other semi-supervised methods with the same baseline, especially regarding E2E results. Similarly, SemiETS also outperforms them on IC15, but the improvements are smaller due to a larger domain shift.
It indicates the promising potential of SemiETS to utilize unlabeled data further. Notably, recent generalist VLLMs lack of text spotting ability, indicating the value of our work in the era of VLLMs.

\begin{table*}[tb]
\scriptsize
\setlength{\tabcolsep}{2.5 mm}
  \caption{Results under Fully Labeled Data setting on Total-Text and ICDAR 2015.}
  \vspace{-8pt}
  \label{tab:full}
  \centering
  \begin{tabular}{llccc|cccc}
    \toprule
    \multirow{3}{*}{Paradigms} & \multirow{3}{*}{Methods} & \multicolumn{3}{c}{Total-Text} & \multicolumn{4}{c}{ICDAR2015} \\\cmidrule(lr){3-5}\cmidrule(lr){6-9}
    &  & Det-F1 &  None & Full & Det-F1 & S & W & G \\ 
    \midrule
    \multirow{7}{*}{\emph{Supervised}} &
    ABCNet v2~\cite{abcnet_v2} & 87.0 & 70.4 & 78.1 & 88.1 & 82.7 & 78.5 & 73.0 \\
    & GLASS~\cite{ronen2022glass} & 88.1 & 79.9 & 86.2 & 85.7 & 84.7 & 80.1 & 76.3  \\
    & TESTR~\cite{testr} & 86.9 & 73.3 & 83.9 & 90.0 & 85.2 & 79.4 & 73.6 \\
    & SwinTextSpotter~\cite{swintextspotter} & 88.0 & 74.3 & 84.1 & - & 83.9 & 77.3 & 70.5 \\
    & TTS (poly)~\cite{tts} & - & 78.2 & 86.3 & - & 85.2 & 81.7 & 77.4 \\
    & DeepSolo~\cite{deepsolo} & 87.3 & 79.7 & 87.0 & 90.0 & 86.8 & 81.9 & 76.9 \\
    & ESTextSpotter~\cite{estextspotter} & 90.0 & 80.8 & 87.1 & 91.0 & 87.5 & 83.0 & 78.1 \\
    \midrule
    \multirow{2}{*}{\emph{VLLMs}} &
    InternVL2-8B~\cite{internvl} & 0.3 & 0.0 & 0.1 & 0.1 & 0.0 & 0.0 & 0.0 \\
    & Qwen2VL2-VL-7B~\cite{qwen} & 1.8 & 0.6 & 1.4 & 16.33 & 15.6 & 14.4 & 13.3 \\
    \midrule
    \multirow{7}{*}{\makecell[l]{\emph{Semi-}\\\emph{supervised}}} &
    Supervised baseline & \underline{87.3} & 79.7 & 87.0 & \textbf{90.0} & \underline{86.8} & \textbf{81.9} & \underline{76.9}\\
    & Mean-Teacher*~\cite{mean_teacher} & 86.9\dminus{-0.4} & 80.1\dplus{+0.4} & 87.1\dplus{+0.1} & \underline{89.4}\dminus{-0.6} & 86.7\dminus{-0.1} & 81.2\dminus{-0.7} & 76.8\dminus{-0.1} \\
    & Soft-Teacher*~\cite{xu2021end} & 86.1\dminus{-1.2} & \underline{80.6}\dplus{+0.9} & 86.7\dminus{-0.3}  & 87.9\dminus{-2.1} & 86.2\dminus{-0.6} & 78.9\dminus{-3.0} & 74.0\dminus{-2.9} \\
    & Unbiased Teacher v2*~\cite{unbiasedv2} & \textbf{87.5}\dplus{+0.2} & 80.3\dplus{+0.6} & \underline{87.3}\dplus{+0.3} & 88.5\dminus{-1.5} & 85.2\dminus{-1.6} & 79.7\dminus{-2.2} & 75.1\dminus{-1.8}\\
    & Semi-DETR*~\cite{zhang2023semi} & 84.0\dminus{-3.3} & 79.6\dminus{-0.1} & 85.5\dminus{-1.5} & 87.9\dminus{-2.1} & 85.4\dminus{-1.4} & 79.6\dminus{-2.3} & 75.0\dminus{-1.9} \\
    \rowcolor{gray!10}
    \cellcolor{white} & SemiETS (Ours) & \textbf{87.5}\dplus{+0.2} &  \textbf{81.7}\dplus{+2.0} & \textbf{87.6}\dplus{+0.6} & \textbf{90.0}\dplus{+0.0} & \textbf{87.0}\dplus{+0.2} & \underline{81.6}\dminus{-0.3} & \textbf{77.0}\dplus{+0.1} \\

    \bottomrule   					
\end{tabular}
\vspace{-12pt}
\end{table*}

\subsubsection{Generalization of our method}

\noindent\textbf{Generalization to text spotters.} 
We also conduct experiments on ABCNet~\cite{abcnet_v1}, a representative RoI-based text spotter. As shown in Tab.~\ref{tab:total_e2e_abc}, SemiETS also brings significant performance gain to the supervised baseline on different text spotting baselines. The proposed strategies achieve consistent improvements over the baseline SSL framework, demonstrating that SemiETS is compatible with both RoI-based and DETR-based text spotters.

\begin{table}[tb]
\scriptsize
\setlength{\tabcolsep}{0.5mm}
  \caption{The end-to-end spotting results on Total-Text under Partially Labeled Data setting with different text spotters.}
  \vspace{-7pt}
  \label{tab:total_e2e_abc}
  \centering
  \begin{tabular}{@{}llcccccccccc@{}}
    \toprule
    \makecell[c]{\multirow{2}{*}{Spotters}} & \makecell[c]{\multirow{2}{*}{Settings}} &  \multicolumn{2}{c}{0.5\%} &  \multicolumn{2}{c}{1\%} &   \multicolumn{2}{c}{2\%} & \multicolumn{2}{c}{5\%} & \multicolumn{2}{c}{10\%} \\ 
    \cmidrule(lr){3-4}\cmidrule(lr){5-6}\cmidrule(lr){7-8}
    \cmidrule(lr){9-10}\cmidrule(lr){11-12}
     & & None & Full & None & Full & None & Full & None & Full & None & Full \\
     \midrule
     \multirow{3}{*}{ABCNet~\cite{abcnet_v1}} & Supervised & 40.9 & 60.1  & 43.7  & 63.6  & 47.6 & 66.0 & 50.4 & 70.1 & 54.7 & 72.9 \\
   & SSL Baseline & 46.1  & 61.4  & 48.1  & 64.0 & 52.2 & 66.9 & 53.1 & 70.8 & 57.3 & 74.6 \\
   & SemiETS &\textbf{54.2}  &\textbf{71.9}  &\textbf{56.0}  &\textbf{74.4}  & \textbf{57.7} & \textbf{74.8} & \textbf{59.2} & \textbf{76.5} & \textbf{61.8} & \textbf{78.7} \\
    \midrule
   \multirow{3}{*}{DeepSolo~\cite{deepsolo}} & Supervised  & 58.8 & 71.1  & 61.2  & 74.9  & 63.4 & 76.8 & 66.9 & 78.8 & 69.9 & 80.9 \\
   & SSL Baseline & 55.5  & 57.7  & 64.4  & 69.0  & 68.6 & 73.7 & 73.5 & 80.6 & 73.8 & 81.9 \\
   & SemiETS &\textbf{72.0}  &\textbf{78.5}  &\textbf{72.8}  &\textbf{80.6}  & \textbf{73.4} & \textbf{82.2} & \textbf{75.4} & \textbf{83.1} & \textbf{76.4}& \textbf{84.8} \\
  \bottomrule
  \end{tabular}
  \vspace{-7pt}
\end{table}

\noindent\textbf{Domain Adaptation.}
We further study the domain adaptation ability using the training set of Total-Text and IC15 as the labeled and unlabeled data, respectively. 
From Tab.~\ref{tab:cross}, the generalization ability of the supervised baseline is limited. Our SemiETS can effectively adapt the text spotter to new scenarios without extra labeling costs, facilitating detection and E2E results by 17.3\% and 14.4\% (generic lexicon), respectively. Moreover, it surpasses other SSL frameworks by a large margin, showing great practical value.

\begin{table}[tb]

\scriptsize
\setlength{\tabcolsep}{1.0mm}
  \caption{Results of domain adaptation. TT is short for Total-Text.}
  \vspace{-7pt}
  \label{tab:cross}
  \centering
  \begin{tabular}{@{}lcc|ccc|ccc|ccc@{}}
    \toprule
    \multirow{2}{*}{Methods} & \multirow{2}{*}{$D_l$} & \makecell[c]{\multirow{2}{*}{$D_u$}} &  \multicolumn{3}{c}{Detection} &  \multicolumn{3}{c}{E2E} &   \multicolumn{3}{c}{Word Spotting} \\ 
    \cmidrule(lr){4-6}\cmidrule(lr){7-9}\cmidrule(lr){10-12}
     & & &  P & R & F1 & S & W & G & S & W & G \\
     \midrule
   Supervised & - & - & 64.8 & 29.9 & 40.9 & 38.9 & 36.1 & 34.0 & 38.5 & 35.9 & 33.7 \\
   Supervised & TT & - & 71.3 & 63.9 &	67.4 & 66.8 &	61.8 & 57.5 & 66.3 & 62.2 & 57.8 \\
   \midrule
   STAC* & TT & IC15 & 65.0 & 62.7 & 63.8 & 62.5 & 56.5 & 51.0 & 62.2 & 56.7 & 51.0 \\
   Mean-Teacher* & TT & IC15 & 76.1 & 89.5 & 82.3 & 82.2 & 71.7 & 65.2 & 81.7 & 72.5 & 65.7 \\
   Soft Teacher* & TT  & IC15 & 77.1 & 88.0 & 82.2 & \underline{83.4} & \underline{74.7} & \underline{67.8} & \underline{83.0} & \underline{75.7} & \underline{68.5} \\
   UT v2* & TT & IC15 & 69.5 & 81.0 & 74.8 & 76.2 & 67.1 & 61.4 & 75.9 & 68.0 & 62.0 \\
   Semi-DETR* & TT & IC15 & 74.5 & 86.2 & \underline{79.9} & 80.0 & 72.8 & 67.5 & 79.6 & 74.0 & \underline{68.5} \\
   \rowcolor{gray!10}
   SemiETS  & TT & IC15 & 81.7 & 88.0 & \textbf{84.7} & \textbf{84.9} & \textbf{77.8} & \textbf{71.9} & \textbf{84.5} & \textbf{78.8} &	\textbf{72.7}  \\
  \bottomrule
  \end{tabular}
  \vspace{-8pt}
\end{table}

\subsection{Ablation Study}

In this section, we conduct extensive studies to validate our designs. We use DeepSolo~\cite{deepsolo} as the baseline text spotter.

\subsubsection{Effect of each component}

We conduct the experiments using various data proportions on Total-Text, demonstrating the effectiveness of the proposed designs shown in Tab.~\ref{tab:abl_component}.
Each component can provide improvements to the baseline in both detection and recognition.
For example, under 2\% data proportion, PSA significantly improves the detection recall and F1 score by 13.4\% and 8.8\%, respectively. The E2E H-mean increases by 3.8\% (6.7\%) with (without) lexicon, respectively. On this basis, MMS can further bring performance gain, achieving the best result on text detection and spotting. Consistent improvements are verified in 5\% data proportion.
Unless specified, the following experiments are conducted using 2\% labeled data setting on Total-Text.

\begin{table}[tb]
\scriptsize
\setlength{\tabcolsep}{1.2mm}
  \caption{The effects of each component. The experiments are conducted on the Partially Labeled Data setting using Total-Text.}
  \vspace{-7pt}
  \label{tab:abl_component}
  \centering
  \begin{tabular}{cc|ccccc|ccccc}
    \toprule
    \multicolumn{2}{c}{\multirow{2}{*}{Settings}} &  \multicolumn{5}{c}{2\%} & \multicolumn{5}{c}{5\%} \\ \cmidrule(lr){3-7} \cmidrule(lr){8-12}    
    & \multicolumn{1}{c}{} & \multicolumn{3}{c}{Detection} &  \multicolumn{2}{c}{E2E} & \multicolumn{3}{c}{Detection} &  \multicolumn{2}{c}{E2E} \\ 
    \cmidrule(lr){1-2} \cmidrule(lr){3-5} \cmidrule(lr){6-7} \cmidrule(lr){8-10} \cmidrule(lr){11-12}
    PSA & MMS & P & R & F1 & None & Full & P & R & F1 & None & Full \\
    \midrule
     & & 96.2 & 58.2 & 72.5 & 68.6 & 73.7 & 95.0 & 71.4 & 81.5 & 73.5 & 80.6 \\
     \checkmark & & 94.1	& 71.6	 & 81.3  & 72.4 & 80.4 & 93.6 & 76.6 & 84.3 & 74.3 & 83.2 \\
    & \checkmark & 95.2	& 66.4	 & 78.2 & 71.3 & 78.5 & 94.4 & 72.9 & 82.3 & 73.7 & 81.5 \\
    \checkmark & \checkmark & 93.5 & 74.2 & \textbf{82.7}  & \textbf{73.4} & \textbf{82.2} & 94.2 & 77.9 & \textbf{85.2}  & \textbf{75.1} & \textbf{83.7} \\
  \bottomrule   						
\end{tabular}
\vspace{-7pt}
\end{table}

\subsubsection{Analysis on PSA}

We remove MMS to study the details of PSA. As shown in Tab.~\ref{tab:psa}, both $\mathcal{T}_r$ and CC can boost the performance individually, and using both simultaneously can further improve performance, indicating the hierarchical label assignment can reduce noisy recognition labels. Specifically, the detection F1 score and E2E H-mean increase by 5.4\% and 3.5\%, respectively. HM can further enhance model performance, especially in detection recall. Due to the complementary nature of the recognition and detection tasks, both tasks benefit concurrently. Integrating all these designs leads to optimal overall performance, particularly on E2E.

\begin{table}[tb]
\scriptsize
\setlength{\tabcolsep}{2.8mm}
  \caption{Ablation study on Progressive Sample Assignment.}
  \vspace{-7pt}
  \label{tab:psa}
  \centering
  \begin{tabular}{ccc|ccccc}
    \toprule
    \multicolumn{3}{c}{Settings} &  \multicolumn{3}{c}{Detection} &  \multicolumn{2}{c}{E2E} \\ \cmidrule(lr){1-3} \cmidrule(lr){4-6} \cmidrule(lr){7-8}
    HM & $\mathcal{T}_R$ & CC & P & R & F1 & None & Full \\
    \midrule
     & &  & 96.2 & 58.2 & 72.5 & 68.6 & 73.7 \\
     &\checkmark & & 95.7 & 61.4 & 74.8  & 70.1 & 75.8  \\
     & & \checkmark  & 95.8	& 63.4	& 76.3 & 71.8 & 77.3  \\
     & \checkmark & \checkmark  & 95.0 	& 66.1	& 77.9 & 72.1 & 78.1  \\
     \checkmark & & & 93.8 & 71.6 & \underline{81.1} & 70.6 & 79.7  \\
    \midrule
    \checkmark & \checkmark  &  & 93.5	& 70.7	 & 80.5 & \underline{72.2} & 79.8  \\
    \checkmark &  & \checkmark & 93.8	& 71.7 & \textbf{81.3} & 72.0 & \underline{80.3}  \\
    \checkmark & \checkmark & \checkmark & 94.1 & 71.6 & \textbf{81.3}  & \textbf{72.4} & \textbf{80.4}   \\
  \bottomrule    						
\end{tabular}
\vspace{-5pt}
\end{table}

\begin{table}[tb]
\scriptsize
\setlength{\tabcolsep}{3.08mm}
\vspace{-2pt}
\caption{Ablation study on Mutual Mining Strategy.}
\vspace{-7pt}
\label{tab:abl_mms_2}
\centering
\begin{tabular}{cc|ccccc}
\toprule
\multicolumn{2}{c}{Settings} & \multicolumn{3}{c}{Detection} & \multicolumn{2}{c}{E2E} \\
\cmidrule(lr){1-2} \cmidrule(lr){3-5} \cmidrule(lr){6-7}
SCI & CRC & P & R & F1 & None & Full \\
\midrule
     &            & 94.1 & 71.6 & 81.3 & 72.4 & 80.4 \\
\checkmark &       & 93.1 & 73.6 & 82.2 & 73.3 & 81.9 \\
     & \checkmark & 93.4 & 73.7 & 82.4 & 73.3 & 81.9 \\
\checkmark & \checkmark & 93.5 & 74.2 & \textbf{82.7} & \textbf{73.4} & \textbf{82.2} \\
\bottomrule   						
\end{tabular}
\vspace{-10pt}
\end{table}

\begin{table}
    \scriptsize
    \centering
    \caption{Ablation study on the detailed designs in the MMS}
    \vspace{-7pt}
    \label{tab:ablation_mmc}
    \resizebox{1.0\linewidth}{!}{
    
    \begin{subtable}[htbp]{0.45\linewidth}
        \centering
        \scriptsize
        \setlength{\tabcolsep}{0.5mm} 
        \caption{The measurements in SCI.}
        \label{tab:abl_SCI}
        \begin{tabular}{@{}l|ccccc@{}}
            \toprule
            \makecell[l]{\multirow{2}{*}{Settings}} &  \multicolumn{3}{c}{Detection} &  \multicolumn{2}{c}{E2E} \\ \cmidrule(lr){2-4} \cmidrule(lr){5-6}
            & P & R & F1 & None & Full \\
            \midrule
            w/o SCI & 94.1 & 71.6 & 81.3 & 72.4 & 80.4 \\
            IoU & 93.8 & 72.2 & 81.6 & 72.8 & 81.0 \\
            DIoU & 93.1 & 73.6 & \textbf{82.2} & \textbf{73.3} & \textbf{81.9} \\
            \bottomrule                         
        \end{tabular}
    \end{subtable}
    
    \hspace*{5pt}
    \begin{subtable}[htbp]{0.45\linewidth}
        \centering
        \scriptsize
        \setlength{\tabcolsep}{0.5mm} 
        \caption{The measurements in CRC.}
        \label{tab:abl_CRC}
        \begin{tabular}{@{}l|ccccc@{}}
            \toprule
            \makecell[l]{\multirow{2}{*}{Settings}} &  \multicolumn{3}{c}{Detection} &  \multicolumn{2}{c}{E2E} \\ \cmidrule(lr){2-4} \cmidrule(lr){5-6}
            & P & R & F1 & None & Full \\
            \midrule
            w/o CRC & 94.1 & 71.6 & 81.3 & 72.4 & 80.4 \\
            DIoU & 93.3 & 72.2 & 81.4 & 72.9 & 80.8 \\
            Levenshtein & 93.4 & 73.7 & \textbf{82.4} & \textbf{73.3} & \textbf{81.9} \\
            \bottomrule                         
        \end{tabular}
    \end{subtable}
    }
\vspace{-8pt}
\end{table}

\subsubsection{Analysis on Mutual Mining Strategy}

We first discuss two key designs in MMS. PSA is equipped to ensure the quality of pseudo labels. As shown in Tab.~\ref{tab:abl_mms_2}. Both SCI and CRC can boost detection and recognition results while combining them leads to the optimal.

\begin{figure}[tb]
  \centering
  \includegraphics[width=0.95\linewidth]{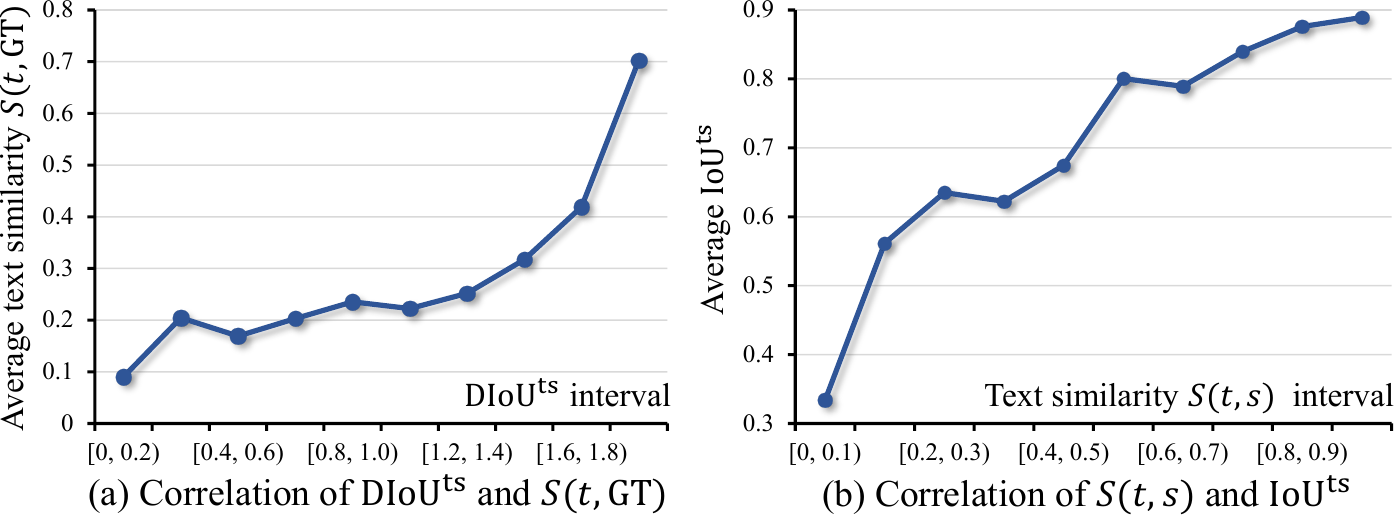}
  \vspace{-5pt}
  \caption{Statistical analysis on the relationship of the accuracy of detected regions and text similarity.}
  \label{fig:correlation}
  \vspace{-5pt}
\end{figure}

\noindent\textbf{Analysis of SCI.} 
The statistical analysis shown in Fig.~\ref{fig:correlation} (a) reveals a strong positive correlation between the region deviation and the average text similarity. This proves that SCI can estimate the reliability of pseudo labels for recognition. Furthermore, we compare the spatial descriptions in Tab.~\ref{tab:abl_SCI}. Our polygon DIoU offers more benefits than vanilla IoU, indicating its advantage in representing text regions.

\noindent\textbf{Analysis of CRC.}
CRC aims to rectify the imprecise text regions with recognition results.
The positive correlation between text similarity and the region alignment measured by IoU shown in Fig.~\ref{fig:correlation} (b) supports our design through statistical analysis.
Yet, another approach is directly using the spatial deviation of the predictions from the teacher and student to assess the quality of regions. We use the normalized DIoU to describe their deviation. Our approach performs better, demonstrating the superiority of mutual mining that effectively boosts task synergy.

\subsection{Visualization Analysis}

\begin{figure}[tp]
  \centering
  \includegraphics[width=1.0\linewidth]{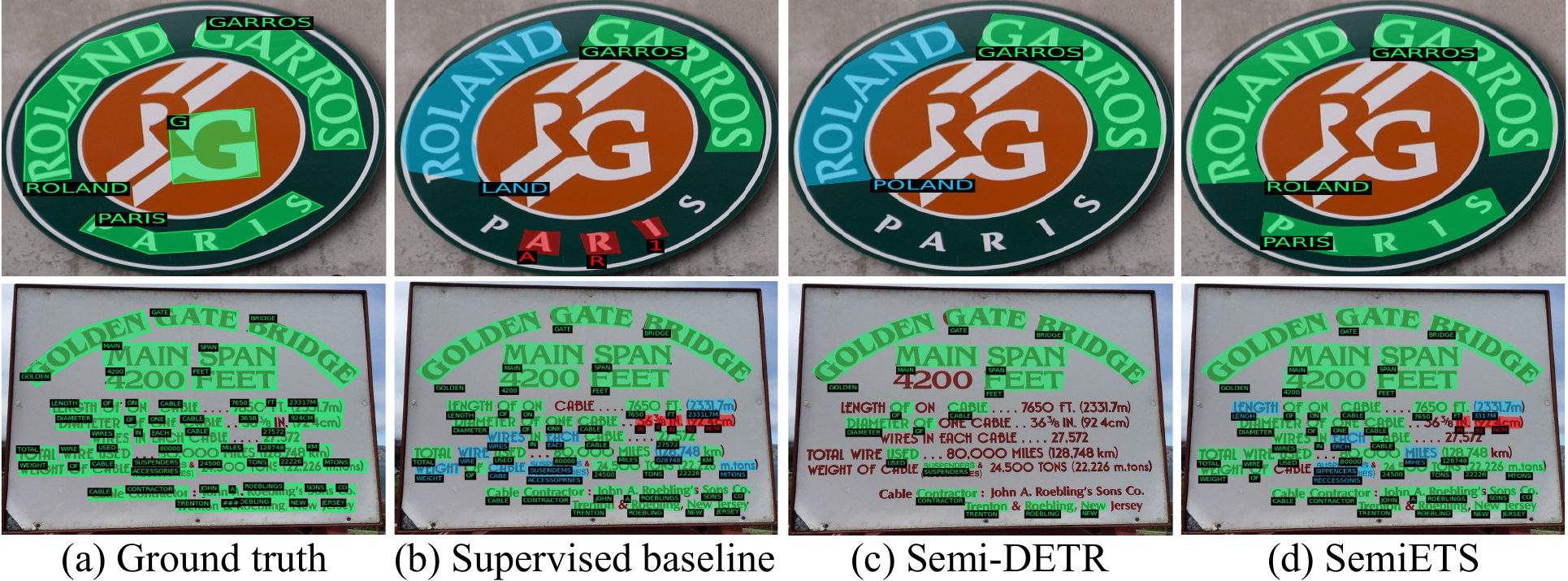}
  \vspace{-15pt}
  \caption{Some typical qualitative results on Total-Text. True positives are indicated in \textcolor{LimeGreen}{green}. Instances in \textcolor{CornflowerBlue}{blue} are detected correctly but recognized wrongly, while \textcolor{BrickRed}{red} are falsely detected.}
  \label{fig:vis}
  \vspace{-15pt}
\end{figure}

\noindent\textbf{Qualitative results.}
We visualize typical qualitative results in Fig.~\ref{fig:vis}.
The first row shows a curved text case, while the second presents a dense text case with varied font sizes. Compared to the supervised baseline and SOTA SSL method, SemiETS achieves higher accuracy and recall with fewer false positives in both scenarios. 
We believe this is because SemiETS extracts richer information from unlabeled data through hierarchical label assignment and inter-task complementarity, highlighting its effectiveness.

\begin{figure}[tb]
  \centering
  \includegraphics[width=1.0\linewidth]{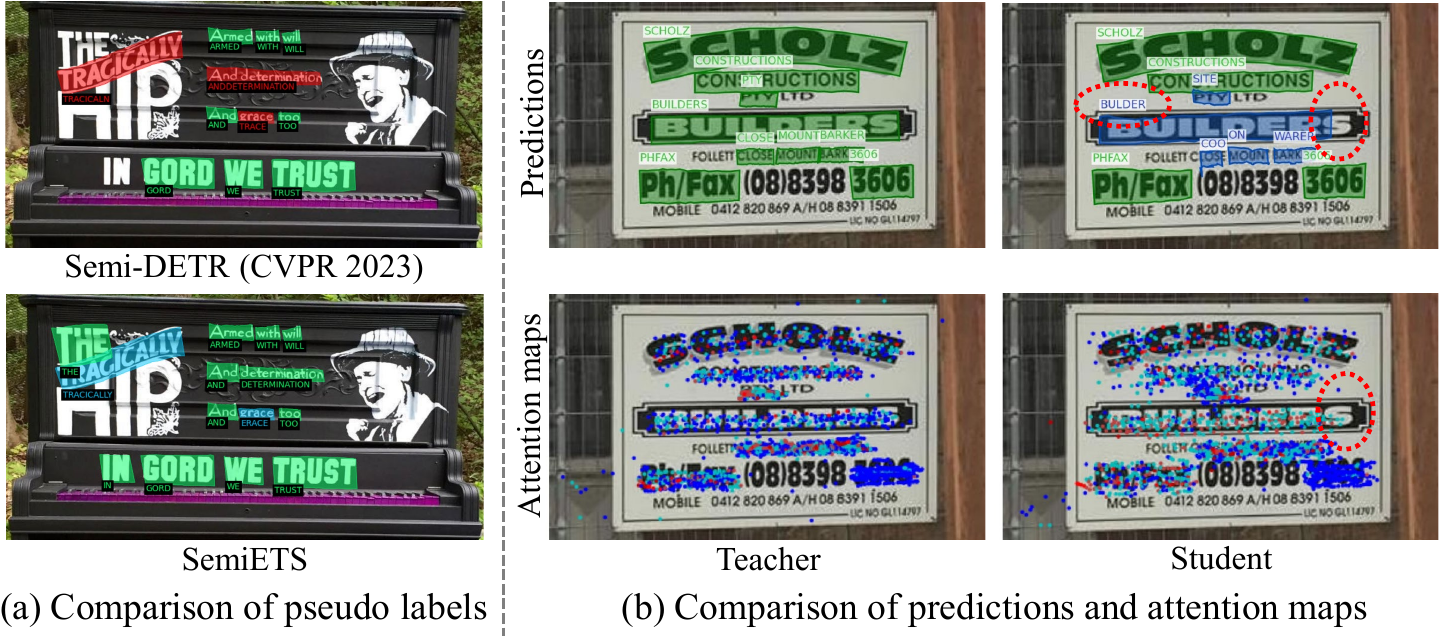}
  \vspace{-15pt}
  \caption{Visualizations of (a) usage of pseudo labels; (b) attention maps of queries in DeepSolo~\cite{deepsolo} and corresponding predictions.}
  \label{fig:attention}
  \vspace{-10pt}
\end{figure}

\noindent\textbf{Hierarchical \emph{v.s.} unitary labels.}
We visualize the used pseudo labels in Fig.~\ref{fig:attention} (a). In Semi-DETR, the simple label assignment approach introduces noisy recognition supervisions (red), which misleads the recognition flow.
In contrast, SemiETS progressively differentiates between Det-only labels (blue) and E2E labels (green), reducing noise and enabling a more reliable label assignment.

\noindent\textbf{Implicit correlation.}
Although the detection and recognition do not have an explicit sequential order in DETR-based spotters, visualization analysis in Fig.~\ref{fig:attention} (b) still supports this property. Most detected regions are highly consistent with their attentions that directly influence the recognition predictions, such as regions in red-dashed circles.

\subsection{Limitations}

Although our method achieves promising results, the usage of the characteristics of texts and the synergy of text detection and recognition is to be explored more deeply.
At present, SemiETS mainly focuses on spotting Latin text. For our future work, we will further explore semi-supervised text spotting in multilingual scenarios.

\section{Conclusion}

The paper presents a straightforward yet effective framework named SemiETS for semi-supervised end-to-end text spotting. Focusing on the challenges of inconsistent pseudo labels, we customize the Progressive Sample Assignment module and Mutual Mining Strategy.
The former enhances the quality of pseudo labels for text spotting.
The latter introduces a crossover strategy to excavate information using the complementarity of text detection and recognition. Compared with the baseline, SemiETS obtains a remarkable improvement and outperforms existing SSOD approaches on several datasets, with extensive experiments demonstrating its generalization and scalability.

\section*{Acknowledgments}

This work was supported by NSFC (Grant No.62225603, 62206104).

{
    \small
    \bibliographystyle{ieeenat_fullname}
    \bibliography{main_arxiv}
}

\clearpage
\setcounter{page}{1}
\maketitlesupplementary

\appendix

\section{Additional Experimental Results}

\subsection{Text Detection Results}

\begin{table}[t]
\scriptsize
\setlength{\tabcolsep} {1.0mm}
  \caption{The text detection results on Total-Text and ICDAR 2015 under the Partially Labeled Data setting. DeepSolo~\cite{deepsolo} is the baseline text spotter consistent with the main experiment.}
  \label{tab:det}
  \centering
  \begin{tabular}{@{}lccccc|ccccc@{}}
    \toprule
    \multirow{3}{*}{Methods} & \multicolumn{5}{c}{Total-Text} & \multicolumn{5}{c}{ICDAR 2015} \\
    \cmidrule(lr){2-6}\cmidrule(lr){7-11}
    & 0.5\% & 1\% & 2\% & 5\% & 10\% & 0.5\% & 1\% & 2\% & 5\% & 10\% \\
     \midrule
   Supervised &  77.1 & 80.2 & \underline{81.4} &  81.8 &  83.6  &  75.5 & 74.9 & 76.7 &  81.4 &  82.6   \\
   \midrule
   STAC* &  \underline{77.8}  &  \underline{80.4} &  80.6 &  \underline{82.8} &  \underline{84.8} &  78.4  &  \underline{79.7} &  81.0 &  \underline{84.0} &  84.2  \\
   Mean-Teacher* & 54.4 & 67.3 &72.5 &  81.5  &  83.7 & 72.1 & 71.2 & 74.8 &  81.3  &  82.6  \\
   Soft Teacher* &  59.2 &  61.8 &  73.6 &  78.9 & 81.3 &  70.2 &  73.4 &  75.4 &  80.9 & 80.2 \\
   UT v2* &  56.0 &  67.3 &  76.7 &  81.4 &  84.5 &  69.3 &  69.2 &  73.3 &  79.0 &  80.3 \\
   Semi-DETR* & 60.6 & 71.8 & 74.8 &  79.8 &  82.0 & \underline{78.9} & 79.6 & \underline{82.2} &  83.6 &  \underline{84.4} \\
   \rowcolor{gray!10}
   SemiETS (Ours) & \textbf{78.8}  & \textbf{80.8}& \textbf{82.7}  & \textbf{84.5} & \textbf{85.4}  & \textbf{80.2}  & \textbf{82.9}& \textbf{83.4}  & \textbf{85.8} & \textbf{86.1}  \\
    
  \bottomrule
  \end{tabular}
\end{table}

As shown in Tab.~\ref{tab:det}, the proposed SemiETS achieves state-of-the-art text detection results on arbitrary-shaped and multi-oriented scene text under all proportions.
Nevertheless, the performance of several existing semi-supervised object detection (SSOD) methods even declines, especially in low data proportions. We attribute this to two aspects. Firstly, the irregular shape of texts increases the difficulty of detection. Secondly, the accumulated error caused by noisy pseudo labels disturbs the optimization.
SemiETS reduces noisy pseudo labels using progressive sample assignment and explicitly enhances the complementarity of detection and recognition by mutual mining, thereby facilitating the performance of both tasks.

\subsection{Additional Domain Adaptation Results}

To simulate diverse domain shifts, We add domain adaptation settings, \ie, from IC15 to Total-Text and from Total-Text to TextOCR. Results in Tab.~\ref{tab:more_da} further demonstrate the consistent improvements in domain adaptation of SemiETS.

\begin{table}[h]
\scriptsize
\setlength{\tabcolsep}{1.0mm}

  \caption{Results of additional domain adaptation experiment (IC15 $\rightarrow$ Total-Text; Total-Text $\rightarrow$ TextOCR).}
  \label{tab:more_da}
  \centering
  \begin{tabular}{@{}l|ccccc|cccc@{}}
    \toprule
    Methods & $D_l$ & $D_u$ & Det-F1 & None & Full & $D_l$ & $D_u$ & Det-F1 & None\\
     \midrule
   Supervised & - & -  & 44.6 & 34.9 & 40.6 & - & - & 32.3 & 22.5 \\
   Supervised & IC15 & -  & 72.9 & 65.0 & 75.8 & TT & - & \underline{54.6} & \underline{41.2} \\
   \midrule
   STAC* & IC15 & TT  & 73.8 & 67.3 & 75.3 & TT & TextOCR & 53.2 & 37.0 \\
   Mean-Teacher* & IC15 & TT  & \underline{76.1} & \underline{69.0} & \underline{77.9} & TT & TextOCR & 52.6 & 33.0 \\
   Soft Teacher* & IC15 & TT  & 68.4 & 64.4 & 71.6 & TT & TextOCR & 47.4 & 33.0 \\
   UT v2* & IC15 & TT & 72.0 & 68.0 & 75.3 & TT & TextOCR & 48.6 & 26.1 \\
   Semi-DETR* & IC15 & TT & 61.9 & 55.9 & 63.5 & TT & TextOCR & 38.1 & 30.8 \\
   \rowcolor{gray!10}
   SemiETS  & IC15 & TT & \textbf{78.6} & \textbf{71.5} & \textbf{80.0} & TT & TextOCR & \textbf{55.3} & \textbf{43.4} \\
  \bottomrule
  \end{tabular}
\end{table}

\subsection{Comparison to VLLMs}

Since generalist vision-language large models (VLLMs) have shown promising performance on various tasks recently, we select recent representative open-source VLLMs, \emph{i.e.,} InternVL2~\cite{internvl} and Qwen2-VL~\cite{qwen}, to verify their effectiveness on our task.
However, results in Fig.~\ref{tab:generator} reveals their limitations in text spotting.
Firstly, as competitive baselines, their spotting results are unsatisfactory.
Secondly, we use them as pseudo-label generators to generate pseudo labels on unlabeled data and then train spotters. Results are even worse as low-quality labels dominate the optimization to the false direction.
It is because VLLMs are good at understanding tasks but are unsuitable for fine-grained perception tasks, indicating the value of our work in the era of VLLMs.

\begin{table}
\scriptsize
\setlength{\tabcolsep}{2.0mm}
  \caption{Comparison to using VLLMs as zero-shot text spotters or label generators using 2\% labeled data on Total-Text.}
  \label{tab:generator}
  \centering
  \begin{tabular}{@{}ll|c|cc@{}}
    \toprule
     Settings & Methods & Det-F1 & None & Full \\
     \midrule
     \multirow{2}{*}{\emph{Zero-shot}} & InternVL2-8B & 0.3 & 0.0 & 0.1 \\
      & Qwen2-VL-7B & 1.8 & 0.6 & 1.4 \\
     \midrule
     \multirow{3}{*}{\emph{Label Generator}} &  InternVL2-8B & 0.0 & 0.0 & 0.0 \\
     & Qwen2-VL-7B & 1.2 & 1.1 & 1.2 \\
   \rowcolor{gray!10}
   \cellcolor{white} & SemiETS  & \textbf{82.7} & \textbf{73.4} & \textbf{82.2}  \\
  \bottomrule
  \end{tabular}
\end{table}

\section{Extensive Ablation Experiments}

\begin{table}[tb]
    \centering
    \scriptsize
    \setlength{\tabcolsep}{2.0mm} 
    \caption{Ablation study on the training stages of applying MMS using 2\% labeled data setting.}
    \label{tab:abl_stage}
    \begin{tabular}{ccc|ccccc}
    \toprule
    \multirow{3}{*}{Settings} & \multicolumn{2}{c}{Applied stages} &  \multicolumn{3}{c}{Detection} &  \multicolumn{2}{c}{E2E} \\ \cmidrule(lr){2-3} \cmidrule(lr){4-6} \cmidrule(lr){7-8}
     & O2M & O2O & P & R & F1 & None & Full \\
    \midrule
    w/o MMS &  &  &  94.1 & 71.6 & 81.3 & 72.4 & 80.4 \\
    Full & \checkmark & \checkmark & 95.5 & 66.5 & 78.4  & 72.6 & 79.2   \\
    O2O & & \checkmark & 93.5 & 74.2 & \textbf{82.7}  & \textbf{73.4} & \textbf{82.2} \\
  \bottomrule   						
\end{tabular}
\end{table}

\noindent \textbf{Training stages.}
For DETR-based spotters, we introduce the stage-wise hybrid matching strategy~\cite{zhang2023semi} to the assignment of PSA to boost the training efficiency, dividing the training process into one-to-many (O2M) and one-to-one (O2O) stage. As shown in Tab~\ref{tab:abl_stage}, applying the Mutual Mining Strategy (MMS) only during the O2O stage achieves the best detection and text spotting results. However, introducing MMS into the O2M stage would cause a decrease in detection performance due to the restriction of recall. 
In early iterations, the pseudo labels generated by the teacher are usually sparse and less reliable. While exploring the potentially high-quality positive proposals using the O2M matching, low-quality predictions would be introduced simultaneously, which might mislead the focus of MMS.
Therefore, MMS is applied only to the O2O stage to refine the guidance after adequate high-quality proposals can be generated.

\noindent \textbf{Diversity of additional data.}
We further explore various unlabeled data sources in the Fully Labeled Data setting on Total-Text in Tab.~\ref{tab:add}. Improvements demonstrate the robustness of SemiETS to utilize unlabeled data. In particular, higher quality and diversity help handle more complex scenes and text styles, bringing more performance gain.

\begin{table}[tb]
\footnotesize
\setlength{\tabcolsep}{1.0mm}
  \caption{Comparison of different additional data for Total-Text.}
  \label{tab:add}
  \centering
  \begin{tabular}{@{}l|ccc@{}}
    \toprule
    \makecell[c]{Settings}  & Det-F1 &  None & Full  \\ 
     \midrule
     Supervised & \underline{87.3} & 79.7 & 87.0 \\
    ~+ MSRA-TD500 & 86.6 & 80.2 & 86.8 \\
    ~+ COCOText & \underline{87.3} & \underline{80.2} & \underline{87.4} \\
    ~+ TextOCR & \textbf{87.5} & \textbf{81.7} & \textbf{87.6} \\
  \bottomrule
  \end{tabular}
  \vspace{-10pt}
\end{table}

\begin{figure}[t]
  \centering
  \includegraphics[width=0.95\linewidth]{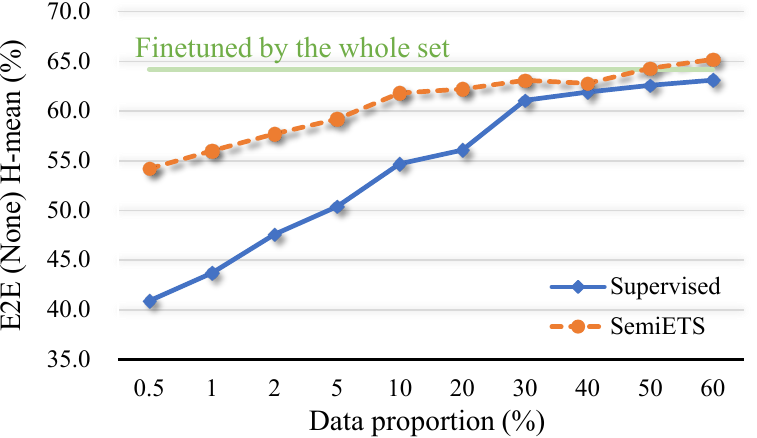}
  \caption{The E2E (None) performance trend of ABCNet on Total-Text under the Partially Labeled Data setting. The \textcolor{LimeGreen}{green} indicates the model finetuned using the whole annotated training set~\cite{abcnet_v1}.}
  \label{fig:trend_abc}
\end{figure}

\noindent \textbf{Parameter study.}
We study the influence of hyper-parameters in Tab.~\ref{tab:para}.
We empirically choose $\mathcal{T}_R = 0.7$ and $\lambda = 20$ by default.

\begin{table}
    \small
    \centering
    \caption{Parameter study.}
    \vspace{-7pt}
    \label{tab:para}
       \resizebox{1.0\linewidth}{!}{
    
    \begin{subtable}[htbp]{0.45\linewidth}
    \centering
    \scriptsize
    \setlength{\tabcolsep}{0.5mm} 
    \caption{The threshold $\mathcal{T}_R$.}
  \label{tab:thres_rec}
  \centering
  \begin{tabular}{@{}l|ccccc@{}}
    \toprule
    $\mathcal{T}_R$   & 0.5 & 0.6 & 0.7 & 0.8 & 0.9 \\\midrule
    Det-F1    & 82.0 & 82.6 & \textbf{82.7} & 82.5 & 81.8  \\
    E2E (None) & 72.4 & 72.6 & \textbf{73.4} & 72.9 & 73.2 \\
    E2E (Full) & 81.0 & 81.3 & \textbf{82.2} & 81.4 & 81.0 \\
  \bottomrule
  \end{tabular}

    \end{subtable}

    \hspace*{5pt}
   \begin{subtable}[htbp]{0.45\linewidth} 
        \centering
        \scriptsize
        \setlength{\tabcolsep}{0.5mm} 
        \caption{The scale factor $\lambda$.}
  \label{tab:lambda}
  \centering
  \begin{tabular}{@{}l|ccccc@{}}
    \toprule
    $\lambda$   & 1 & 10 & 20 & 50 & 100 \\\midrule
    Det-F1 & 81.7 & 82.3 & \textbf{82.7} & 82.3 & 82.5  \\
    E2E (None) & 73.0 & 73.1 & \textbf{73.4} & 73.1 & 73.1 \\
    E2E (Full) & 81.6 & 81.2 & \textbf{82.2} & 81.4 & 81.5 \\
  \bottomrule
  \end{tabular}
    \end{subtable}
}
\end{table}

\section{Performance Trend}

We gradually increase the proportion of labeled data of Total-Text under the Partially Labeled Data setting and display the performance trend of ABCNet~\cite{abcnet_v1} on E2E H-mean without lexicon in Fig.~\ref{fig:trend_abc}.
SemiETS can significantly boost text spotting performance compared to the supervised baseline, and the improvement is more notable when using less labeled data. Furthermore, as the proportion of annotated data increases, E2E H-mean continues growing. When only using 50\% labeled data, SemiETS even outperforms the model finetuned using the whole labeled training set of Total-Text referred from~\cite{abcnet_v1}, demonstrating the potential of the proposed framework to effectively reduce labeling cost and explore useful information from unlabeled data.

\section{More Visualization Results}

\subsection{Pseudo Labels}

\begin{figure}[tb]
  \centering
  \includegraphics[width=1.0\linewidth]{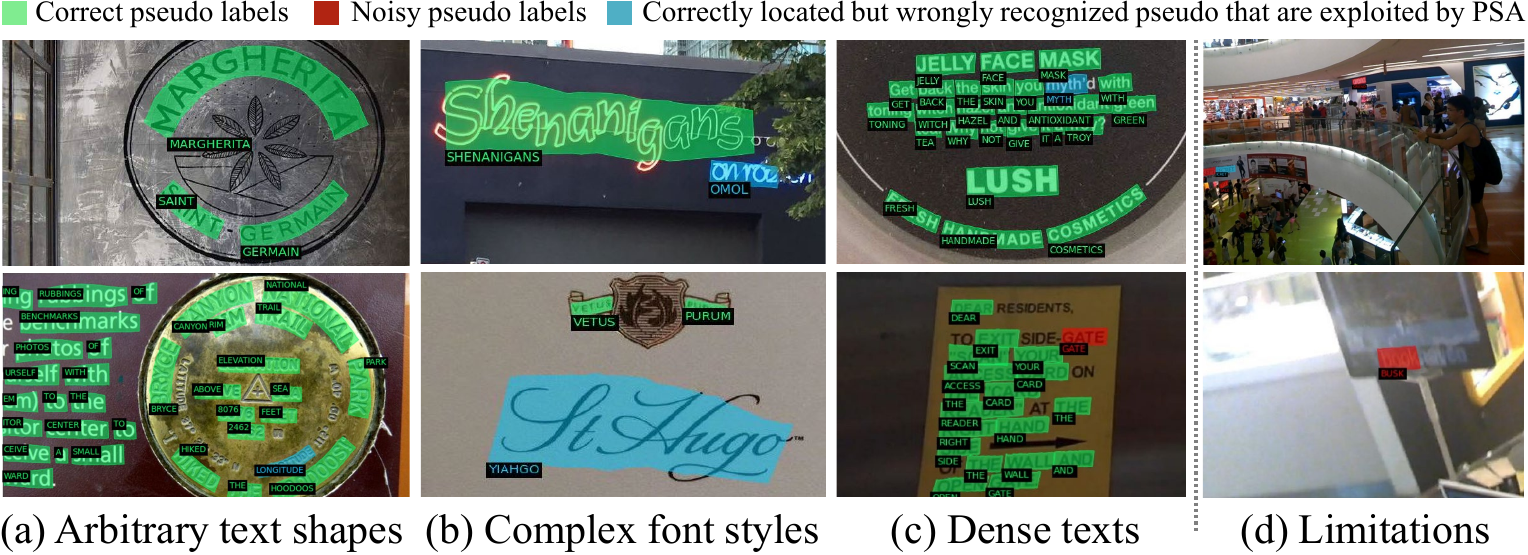}
   \caption{Visualization of pseudo labels generated by SemiETS in typical scenarios.}
  \label{fig:text_cases}
\end{figure}

We visualize pseudo labels generated by SemiETS in several challenging scenarios shown in Fig.~\ref{fig:text_cases} to examine its effectiveness and potential limitations.
1) Arbitrary-shaped texts increase the difficulty of obtaining precise localization labels. SemiETS can handle them with the proposed MMS to rectify text location.
2) Complex text fonts would lead to incorrect pseudo recognition labels.
SemiETS can distinguish them and alleviate noisy recognition labels while still making use of reliable localization labels with the proposed PSA.
3) Dense texts would lead to label omission or shift due to adjacent interference. SemiETS exhibits decent pseudo label generation ability to some extent, as it imposes fine-grained constraints.
However, for some extremely tiny and blurry texts, SemiETS still faces challenges.

\begin{figure*}[tb]
  \centering
  \includegraphics[width=1.0\linewidth]{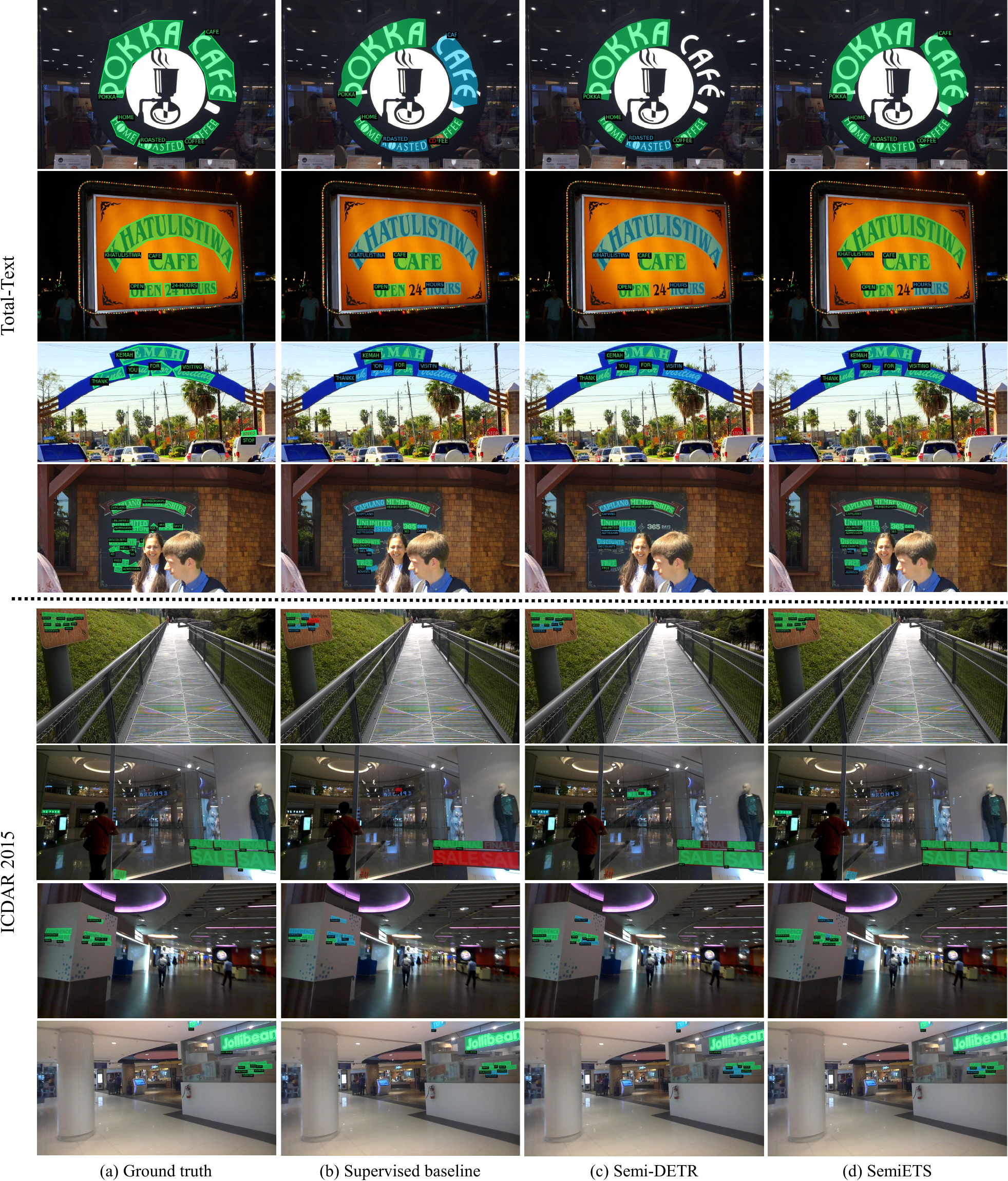}
  \caption{Qualitative results on Total-Text and ICDAR 2015. True positives are indicated in \textcolor{LimeGreen}{green}. Text instances in \textcolor{CornflowerBlue}{blue} are localized accurately but recognized incorrectly. Instances in \textcolor{BrickRed}{red} are inaccurately localized.}
  \label{fig:more_vis}
\end{figure*}

\subsection{Qualitative Results}

We visualize representative qualitative results from Total-Text~\cite{total} and ICDAR 2015~\cite{ic15} in Fig.~\ref{fig:more_vis}. SemiETS demonstrates superior performance in detecting and localizing curved and multi-oriented scene texts while significantly minimizing recognition errors. This improvement stems from its progressive sample assignment mechanism, effectively mitigating noisy supervision signals for text recognition, and its mutual mining strategy, which aims at extracting important guidance information. The robustness of SemiETS gets further validated in challenging scenarios, including incidental and densely distributed scene texts.

\end{document}